\documentclass[10pt,twocolumn,letterpaper]{article}

\usepackage[pagenumbers]{cvpr} % Preprint (arXiv) version: author block, page numbers, no review ruler

\usepackage{amsmath,amssymb,booktabs,multirow,graphicx,xcolor}
\usepackage{array}
\usepackage{colortbl}
\IfFileExists{fontawesome5.sty}{\usepackage{fontawesome5}}{\newcommand{\faGithub}{}}
\definecolor{pBg}{HTML}{F7FAFC}     % panel background
\definecolor{pSky}{HTML}{A7C7E7}    % ours, intermediate (step 2 / camera half)
\definecolor{pIce}{HTML}{CFEAF6}    % ours, base
\definecolor{pCoral}{HTML}{F4A7AE}  % failure / strongest baseline / on-policy shift
\definecolor{pPeach}{HTML}{FBD9C9}  % other baselines / initial-state half
\definecolor{pNavy}{HTML}{5878A6}   % LoopQ (final), emphasis
\definecolor{pMint}{HTML}{A6D0C6}   % step 1: closed-loop calibration
\definecolor{pGrey}{HTML}{DADDE3}   % full precision, grids
\definecolor{pInk}{HTML}{2E3F66}    % text on figures

\definecolor{cvprblue}{HTML}{5878A6}
\usepackage[pagebackref,breaklinks,colorlinks,allcolors=cvprblue]{hyperref}

\def\confName{CVPR}
\def\confYear{2027}

\newcommand{\method}{LoopQ}
\newcommand{\pizero}{$\pi_{0.5}$}
\definecolor{pCoralDark}{HTML}{D9737D}
\newcommand{\tabstyle}{\setlength{\aboverulesep}{0pt}\setlength{\belowrulesep}{0pt}\renewcommand{\arraystretch}{1.17}}
\newcommand{\hdrrow}{\rowcolor{pGrey!55}}           % header band (neutral grey)
\newcommand{\grprow}{\rowcolor{pPeach!42}}          % group label row (peach)
\newcommand{\calrow}{\rowcolor{pMint!42}}           % closed-loop calibration only (mint, step 1)
\newcommand{\oursrow}{\rowcolor{pIce!75}}           % LoopQ (blue)
\newcommand{\dn}[1]{\textcolor{pCoralDark}{$-$#1}}  % points below full precision
\newcommand{\Hcl}{H^{\mathrm{cl}}}
\newcommand{\tr}{\operatorname{tr}}

\title{Quantizing in the Loop:\\Drift-Aware Ultra-Low-Bit Vision-Language-Action Models}

\newcommand{\codeurl}{https://github.com/ForeverBlue816/LoopQ}
\author{Yanlong Chen \quad Thongthornpatch Chamadol \quad Yichen Guo \quad Yunfei Gao\\
Mohan Liu \quad Addison Lin Wang \quad Yawei Li$^{\dagger}$\\[3pt]
Nanyang Technological University\\[1pt]
{\small $^{\dagger}$Corresponding author}\\[3pt]
{\normalsize\faGithub~\href{\codeurl}{\texttt{github.com/ForeverBlue816/LoopQ}}}}

\begin{document}
\maketitle

%=====================================================================
\begin{abstract}
Deploying large vision-language-action (VLA) models on resource-constrained robots demands extreme compression, yet existing post-training quantization (PTQ) methods frequently collapse at ultra-low bit-widths ($\le$ 2 bits). We trace this failure to two mismatches in standard pipelines. First, calibrating on static offline demonstrations induces \textbf{distribution shift}: minor quantization perturbations drive the policy into unseen states where uncalibrated errors compound. Second, minimizing the layer-wise weight reconstruction error is a \textbf{flawed proxy}, as action generation depends only on the representations consumed by the action head. We present \textbf{\method{}}, an on-policy ultra-low-bit quantization framework. \method{} \textbf{closes the loop} by calibrating on states from the quantized policy's own rollouts. It then adds \textbf{readout-aligned fine-tuning}, which freezes the trellis codes and optimizes only lightweight lookup tables and scaling factors to align the backbone readout with the full-precision teacher. With all linear layers at 2 bits, \method{} achieves a \textbf{60.4\%} success rate on LIBERO-Plus with GR00T~N1.7, close to the 61.8\% of full precision. Even at \textbf{1 bit} per weight, \pizero{} still completes \textbf{80.0\%} of LIBERO tasks. With 8-bit activations, the 2-bit \pizero{} runs on integer kernels on an embedded \textbf{NVIDIA Jetson AGX Thor}, with \textbf{4.1$\times$} smaller weights and \textbf{44\%} less memory. On a real \textbf{AgileX Piper} arm, it stays within one success of full precision on each of three tasks. Code and checkpoints are available at \faGithub~\href{\codeurl}{\texttt{github.com/ForeverBlue816/LoopQ}}.
\end{abstract}

%=====================================================================
\begin{figure*}[t]
\centering
\includegraphics[width=\linewidth]{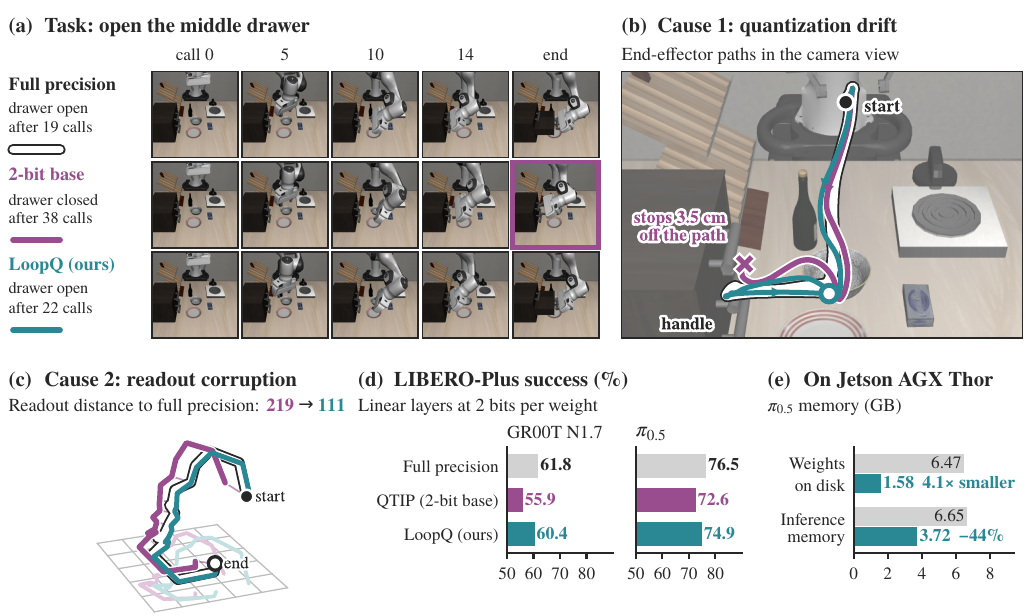}
\caption{\textbf{Closed-loop failure of a 2-bit VLA and its correction by \method{}.} \textbf{(a)}~GR00T~N1.7 on a LIBERO-Plus instance (Goal suite, perturbed initial state) with identical noise, at matched policy calls. Full precision and \method{} succeed, whereas the 2-bit post-training baseline fails. \textbf{(b)}~\emph{Cause 1: quantization drift.} End-effector trajectories on the initial frame. The 2-bit baseline departs from the full-precision trajectory and halts, while \method{} tracks it to completion. \textbf{(c)}~\emph{Cause 2: readout corruption.} Readouts of the \pizero{} action expert along the rollout, in their first three principal components. \method{} halves the distance to full precision. \textbf{(d)}~LIBERO-Plus success with linear layers at 2 bits per weight. \textbf{(e)}~Storage and memory of \pizero{} with 8-bit activations on Jetson AGX Thor.}
\label{fig:teaser}
\end{figure*}

%=====================================================================
\section{Introduction}
\label{sec:intro}

Vision-language-action (VLA) models~\cite{brohan2023rt,kim2024openvla,black2024pi_0,intelligence2025pi_,bjorck2025gr00t} have become the standard recipe for generalist robot policies: a pretrained vision-language backbone, a flow-matching or diffusion transformer action head~\cite{lipman2022flow,chi2025diffusion}, and several billion parameters evaluated many times per second on onboard hardware. While language models of this scale are routinely compressed to 2 bits per weight with modest loss~\cite{tseng2024quip,egiazarian2024extreme,tseng2024qtip}, VLAs resist such extreme compression. Existing VLA quantizers stop at 4 bits or keep the action head in higher precision~\cite{xu2026qvla,zhang2026quantvla,akbari2026actquant}; applying the strongest 2-bit quantizers across an entire VLA drops perturbed success by 4 to 62 points. \Cref{fig:teaser}a illustrates this failure: from the same initial state and noise, the full-precision policy opens the drawer, whereas its 2-bit counterpart reaches for the upper drawer and stalls.

Why does a recipe that works for language models break a policy? Because \emph{a policy acts in closed loop}: its outputs determine its next inputs. This single fact invalidates two core assumptions behind standard post-training quantizers, explaining both why they fail and how to fix them.

\textbf{Quantization drift.} Layer-wise quantizers minimize output error on static calibration data (human demonstration frames), pushing the unavoidable ultra-low-bit error into input directions that those frames never excite. While a language model rarely encounters this hidden error because its test and calibration distributions align, a closed-loop robot inevitably does: a 2-bit policy acts slightly differently from full precision, enters an unseen state, triggers the error hidden there, and deviates further. We term this self-reinforcing divergence \emph{quantization drift}. \Cref{fig:teaser}b traces it in a real episode, where the 2-bit end-effector leaves the reference path and never recovers. Across every VLA module, policy-induced states dominate the input distribution shift at deployment, accounting for 45\% in the language model and up to 83\% in the action head.

\textbf{Readout corruption.} Because the action head interacts with the backbone solely through attention readouts over backbone tokens, action generation depends not on raw weights, but on what the head reads. Ultra-low-bit quantization systematically displaces this interface, shifting the readout by a nearly constant offset at every step of an episode (\cref{fig:teaser}c). The head is not reading a noisier copy of the same content; it is reading something else.

\method{} translates both diagnoses into two steps atop a trellis-coded format~\cite{marcellin1990trellis,tseng2024qtip}. \emph{Closed-loop calibration} quantizes every layer on the states the quantized policy itself visits, so the quantizer can no longer hide error where the policy will go; it requires a few hundred rollouts and no training. \emph{Readout-aligned fine-tuning} then updates only a 512-entry lookup table and a scale vector per layer around the frozen codes, aligning the action head's readouts with the full-precision model. Neither step alters the storage format, bit budget, or decoder.
The result is a VLA quantized end to end, across both backbone and action head, that behaves like the original (\cref{fig:teaser}d). Without any training, closed-loop calibration alone outperforms the strongest post-training quantizer on LIBERO-Plus, reaching 73.4\% vs.\ 72.6\% on \pizero{} and 57.5\% vs.\ 55.9\% on GR00T~N1.7. Adding readout alignment brings the 2-bit policies to 74.9\% and 60.4\%, within 1.6 and 1.4 points of full precision, recovering three quarters of the degradation incurred by the best existing 2-bit quantizer on GR00T. Nor does our approach stop at 2 bits: at \emph{1 bit} per weight, \pizero{} still completes 80.0\% of LIBERO, far surpassing the 48.0\% reported for 2-bit backbone-only quantization~\cite{akbari2026actquant}. Finally, the compressed policy is directly deployable: with 8-bit activations, the 2-bit \pizero{} executes on integer kernels on an embedded NVIDIA Jetson AGX Thor with $4.1\times$ smaller weights and 44\% less memory (\cref{fig:teaser}e). On a real AgileX Piper arm, the 2-bit policy succeeds in 58 of 75 trials, against 59 for full precision. Our contributions are:
\begin{itemize}
\item \textbf{A closed-loop diagnosis.} We show that ultra-low-bit VLAs fail due to two phenomena rooted in closed-loop execution: \emph{quantization drift}, caused by calibration statistics that miss policy-induced states, and \emph{readout corruption}, a systematic displacement of the representations consumed by the action head under layer-wise objectives.
\item \textbf{\method{}.} We propose \emph{closed-loop calibration} to align calibration statistics with the policy's own state distribution, outperforming the strongest post-training quantizers on two VLAs without any training, followed by \emph{readout-aligned fine-tuning}, which optimizes only per-layer lookup tables and scales while keeping all stored codes and the decoder untouched.
\item \textbf{End-to-end VLAs at 2 bits and below.} We present the first VLAs with both backbone and action head quantized entirely to 2 bits that stay within two points of full precision under perturbation, while remaining functional policies even at 1 bit. Furthermore, we deploy the 2-bit policy on a physical AgileX Piper arm, demonstrating reliable closed-loop execution across three real-world manipulation tasks: precision stacking, two-stage lidded-bin lemon placement, and multi-object grasping in clutter, each within one success of full precision.
\end{itemize}

%=====================================================================
\section{Related Work}
\label{sec:related}

\textbf{Vision-language-action models.} RT-2~\cite{brohan2023rt} and OpenVLA~\cite{kim2024openvla} cast actions as tokens of a vision-language model; $\pi_0$ and \pizero{}~\cite{black2024pi_0,intelligence2025pi_} and the GR00T family~\cite{bjorck2025gr00t} attach a flow-matching action head that reads the backbone through attention and denoises an action chunk in a few steps. We study the latter family, where the interface between backbone and head is an attention readout.

\textbf{Low-bit quantization of large models.} GPTQ~\cite{frantar2022gptq} and AWQ~\cite{lin2024awq} quantize layer by layer using second-order statistics of the layer inputs, following the optimal-brain line of work~\cite{hassibi1992second}. At two bits, scalar quantizers collapse and vector formats take over: QuIP and QuIP\#~\cite{chee2023quip,tseng2024quip} combine incoherence processing with lattice codebooks, AQLM~\cite{egiazarian2024extreme} and VPTQ~\cite{liu2024vptq} learn additive and vector codebooks, and QTIP~\cite{tseng2024qtip} uses trellis-coded quantization~\cite{marcellin1990trellis} to approach the rate--distortion limit with a decoder that needs only a small lookup table. Rotations reshape weight and activation statistics before quantization~\cite{ashkboos2024quarot,liu2025spinquant,chen2026prismquant}, and fine-tuning or distillation repairs the continuous parameters afterwards~\cite{tseng2024quip,egiazarian2024extreme,malinovskii2024pv,chen2026gated}. We adopt the trellis-coded format and contribute what is specific to policies: the calibration statistics and the fine-tuning objective.

\textbf{Quantizing VLAs.} QVLA~\cite{xu2026qvla}, QuantVLA~\cite{zhang2026quantvla}, ActQuant~\cite{akbari2026actquant}, and DA-PTQ~\cite{xu2026ptq} adapt bit allocation, scaling, or calibration to action sensitivity, mostly at four bits or with parts of the action pathway kept in higher precision. BitVLA~\cite{wang2025bitvla} instead train the policy with quantization in the loop. VLAQuantBench~\cite{xu2026vlaquantbench} shows that a single action-interface layer can dominate the loss. These works calibrate on demonstrations; we quantize every linear layer to two bits and identify the closed loop as the root cause of failure.

\textbf{Calibration data and distribution shift.} Post-training quantization is sensitive to its calibration set~\cite{williams2024impact}. In imitation learning, DAgger~\cite{ross2011reduction} corrects the mismatch between demonstrations and learner-induced states by querying an expert and retraining. Closed-loop calibration needs neither labels nor training: it re-estimates layer statistics on the states the quantized policy itself visits.
%=====================================================================
\section{Failure Analysis of Low Bits VLAs}
\label{sec:analysis}

\subsection{Preliminaries}
A flow-matching VLA~\cite{black2024pi_0,intelligence2025pi_} consists of a \emph{prefix} network (vision encoder and language model), evaluated once per inference, and an \emph{action head}, evaluated at each of $T$ denoising steps~\cite{lipman2022flow}. At layer $\ell$, step $t$ and attention head $h$, the action head forms the readout
\begin{equation}
o_{\ell,t,h} = \operatorname{softmax}\!\big(q_{\ell,t,h}K_\ell^{\top}/\sqrt{d}\big)\,V_\ell ,
\label{eq:readout}
\end{equation}
where the queries $q$ come from the action tokens and $(K_\ell,V_\ell)$ from the prefix: the cached keys and values in \pizero{}, the backbone tokens read by the cross-attention of the diffusion transformer in GR00T \cite{peebles2023scalable}. The readouts are the only path from perception to action.

Post-training quantizers replace each weight matrix $W_\ell$ by $\hat W_\ell$ that minimizes the layer-wise objective~\cite{frantar2022gptq}
\begin{equation}
\begin{aligned}
&\min_{\hat W_\ell}\ \tr\!\big((\hat W_\ell-W_\ell)\,H_\ell\,(\hat W_\ell-W_\ell)^{\top}\big),\\
&H_\ell=\mathbb{E}_{x\sim\mathcal D}\big[xx^{\top}\big],
\end{aligned}
\label{eq:layerwise}
\end{equation}
with the second moment $H_\ell$ of the layer input estimated on a calibration set $\mathcal D$ of demonstration frames.

\begin{figure*}[t]
\centering
\includegraphics[width=\linewidth]{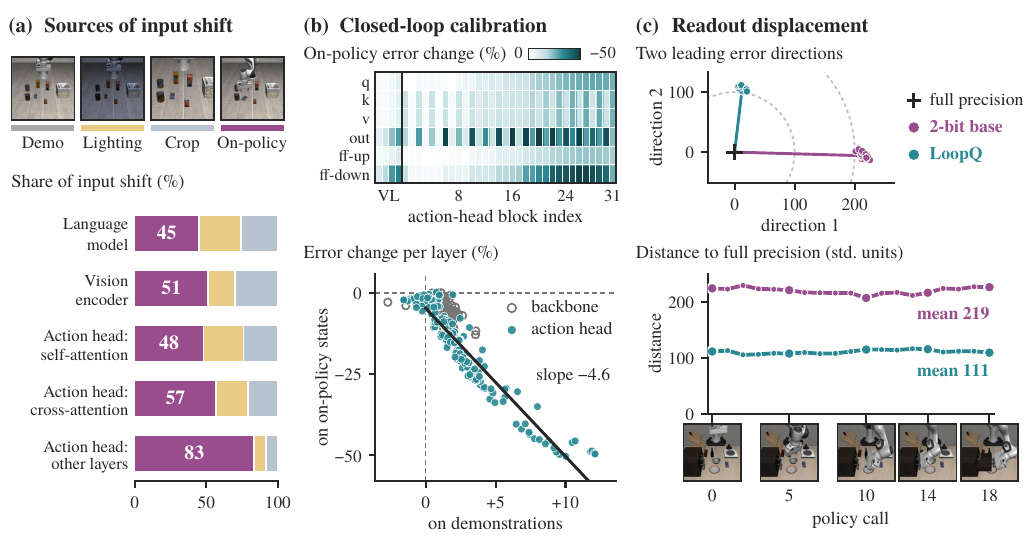}
\caption{\textbf{The two failure causes of 2-bit VLAs and their correction.} \textbf{(a)}~Decomposition of the deployment input shift across modules, indicating that on-policy states outweigh visual perturbations. \textbf{(b)}~Layer-wise effect of closed-loop calibration relative to demonstration calibration. On-policy error decreases across the action head (top) and in 427 of 432 layers. Each 1\% added on demonstrations removes 4.6\% on on-policy states in the head (bottom). \textbf{(c)}~Action-expert readout error on identical inputs from the pictured rollout, in the two leading directions of both models' errors (top) and over policy calls (bottom). \method{} halves the mean distance to full precision.}
\label{fig:phenomena}
\end{figure*}

\subsection{Quantization drift under closed-loop execution}
Equation~\eqref{eq:layerwise} weights the reconstruction error by Hessian calibration $H_\ell$. Because errors along low-energy directions of $H_\ell$ incur negligible penalty, an ultra-low-bit quantizer inevitably hides its unavoidable distortion in those subspaces. For $y=w_1x_1+w_2x_2$, the output error is $\Delta w_1x_1+\Delta w_2x_2$: if $x_2\approx 0$ on the demonstration set $\mathcal D$, $\Delta w_2$ is unconstrained and $w_2$ is quantized coarsely. This is harmless only when $x_2$ is near zero. Because a policy's actions alter the future scene, small quantization perturbations steer a 2-bit policy into states where $x_2\neq0$, exposing the hidden error $\Delta w_2x_2$, amplifying the deviation, and closing a self-reinforcing loop \cite{ross2010efficient}. To quantify this effect, we decompose the second layer-wise moment of the deployment input shift relative to calibration frames across three sources: photometric variations, geometric variations, and states visited by the 2-bit policy itself. Across the entire VLA, policy-induced states dominate the distribution shift, accounting for 45\% in the language model, 51\% in the vision encoder, 48--57\% in the action head's attention layers, and up to 83\% in its remaining layers (\cref{fig:phenomena}a).

To isolate this mismatch, we change only the calibration statistics so that they include on-policy states (\cref{sec:method}) and keep everything else fixed (\cref{fig:phenomena}b). The output error on on-policy states then falls in 427 of the 432 layers, most strongly in the action head (median 14\%, up to 50\%) and least in the backbone (median 3\%). The improvement comes at a nearly linear cost on demonstrations: in the action head, each 1\% of additional error on demonstrations corresponds to a 4.6\% reduction on on-policy states. The gain does not appear along individual principal directions of the layer inputs: on-policy calibration does not reduce the per-direction error, but aligns the errors so that they cancel on the inputs seen at deployment. This indicates that fidelity on demonstrations and fidelity under closed-loop deployment are distinct objectives, and the calibration data decide which one the quantizer optimizes.

\subsection{Readout corruption in the action head}
Calibration statistics determine where weight error is placed, and the second failure cause lies in the layer-wise objective itself, which leaves the readout of the action head unconstrained~\cite{gong2025pushing,kim2025guidedquant}. To measure the effect, we feed the same episode, with identical inputs, to the full-precision \pizero{} and to its 2-bit versions and record the attention readout of the action expert at each of the 19 policy calls (\cref{fig:phenomena}c). Quantization to 2 bits displaces the readout by a nearly constant offset at every policy call. The plot axes are the two leading directions of both models' errors: the base error lies along the first, and \method{} removes it, leaving a residual half as large along the second. \method{} is closer to full precision at all 19 calls (0.51 of the base distance on average). The trellis format and closed-loop calibration bring it to 0.63, and readout alignment accounts for the rest.

%=====================================================================
\section{Method}
\label{sec:method}

\begin{figure*}[t]
\centering
\includegraphics[width=\linewidth]{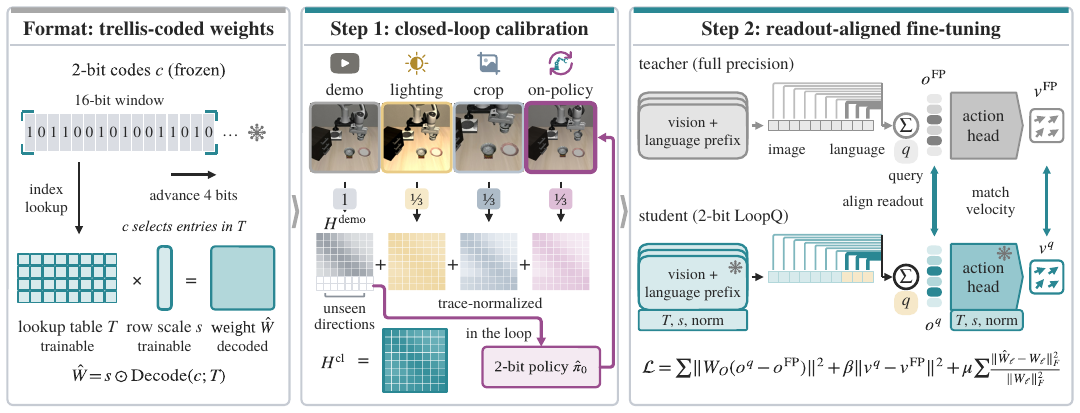}
\caption{\textbf{Overview of \method{}.} \textbf{Format.} A 16-bit window slides over the frozen trellis code stream and indexes a 512-entry table $T$. A row scale $s$ multiplies each output. \textbf{Step 1.} Closed-loop calibration adds trace-normalized lighting, crop and on-policy statistics to $H^{\mathrm{demo}}$. \textbf{Step 2.} Readout-aligned fine-tuning trains only $T$, $s$ and the normalization weights to match the teacher's readout $o$ and velocity $v$.}
\label{fig:method}
\end{figure*}

\method{} keeps the quantization format fixed and changes two things around it: the statistics that guide the quantizer and the objective that repairs its output (\cref{fig:method}).

\subsection{Base format: trellis-coded quantization}
We quantize every linear layer of the prefix and the action head with trellis-coded quantization~\cite{marcellin1990trellis} in the bitshift formulation of QTIP~\cite{tseng2024qtip}. Each weight matrix undergoes a two-sided randomized Hadamard transform and is partitioned into 256-weight tiles. Each tile is encoded as a 512-bit path through a $2^{16}$ state trellis: the decoder slides a 16-bit window over the bitstream in 4-bit strides, hashes the window to index a 512-entry lookup table $T_\ell$, and outputs a weight pair scaled by a per-row factor $s_\ell$. Viterbi search with block-wise error feedback selects the path minimizing \cref{eq:layerwise}. Because adjacent windows overlap, all weights within a tile are optimized jointly, allowing a compact 512-pair table to approach the 2-bit rate--distortion limit~\cite{gray1998quantization}.

\subsection{Closed-loop calibration}
To eliminate the distribution mismatch between static demonstrations $\mathcal D$ and closed-loop execution, the calibration second moment should be estimated under the state distribution $d_{\hat\pi}$ induced by the quantized policy $\hat\pi(H)$ that it produces, which defines a fixed point:
\begin{equation}
H^{\star}_\ell=\mathbb{E}_{x\sim d_{\hat\pi(H^{\star})}}\big[xx^{\top}\big].
\end{equation}
Standard demonstration statistics $H^{\mathrm{demo}}_\ell$ correspond to the zeroth iterate of this fixed-point problem. We perform one fixed-point iteration and incorporate visual invariance by constructing the closed-loop Hessian:
\begin{gather}
\Hcl_\ell = H^{\mathrm{demo}}_\ell+\sum_{s\in\mathcal S} w_{\ell,s}\,H^{(s)}_\ell, \label{eq:hcl} \\
w_{\ell,s}=\frac{\tr H^{\mathrm{demo}}_\ell}{|\mathcal S|\,\tr H^{(s)}_\ell}, \label{eq:hcl_w}
\end{gather}
where $\mathcal S$ comprises three sources: lighting augmentations and random crops of the calibration frames, and \emph{on-policy states}, one from each of 256 rollouts of the 2-bit policy calibrated on demonstrations in the unperturbed environment. Trace normalization gives every source the same total energy. We start these rollouts from the benchmark's full distribution of initial states, which covers more states than a single start pose and preserves initial-state robustness (\cref{sec:ablation}). The perturbations used for evaluation never appear in the calibration data.

\begin{table*}[t]
\centering
\small\tabstyle
\setlength{\tabcolsep}{7.4pt}
\begin{tabular}{l c c c cc cccc}
\toprule
\hdrrow & & & & \multicolumn{2}{c}{By perturbation} & \multicolumn{4}{c}{By suite} \\
\hdrrow Method & bpw & All & vs.\ FP & Camera & Init.\ state & Spatial & Object & Goal & Long \\
\midrule
\grprow\multicolumn{10}{l}{\emph{GR00T~N1.7: all 432 linear layers of backbone and action head at 2 bits}}\\
Full precision & 16 & 61.8 & -- & 65.2 & 58.4 & 74.7 & 50.2 & 69.5 & 52.8 \\
AQLM~\cite{egiazarian2024extreme} & 2.02 & 40.1 & \dn{21.7} & 40.2 & 40.0 & 58.4 & 39.0 & 40.6 & 22.6 \\
VPTQ~\cite{liu2024vptq} & 2.00 & 49.5 & \dn{12.3} & 51.6 & 47.4 & 65.7 & 44.2 & 55.9 & 32.3 \\
QuIP\#~\cite{tseng2024quip} & 2.03 & 53.1 & \dn{8.7} & 54.4 & 51.8 & 67.3 & 45.0 & 59.0 & 41.1 \\
QTIP~\cite{tseng2024qtip} & 2.02 & 55.9 & \dn{5.9} & 54.4 & 57.4 & 71.4 & 48.6 & 60.2 & 43.5 \\
\calrow Closed-loop calibration only (ours) & 2.02 & 57.5 & \dn{4.3} & 57.8 & 57.2 & 71.0 & \textbf{50.6} & 65.2 & 43.1 \\
\oursrow \textbf{\method{}} (ours) & 2.02 & \textbf{60.4} & \textbf{\dn{1.4}} & \textbf{62.2} & \textbf{58.6} & \textbf{75.1} & 49.0 & \textbf{68.4} & \textbf{49.2} \\
\midrule
\grprow\multicolumn{10}{l}{\emph{\pizero{}: all 414 linear layers of backbone and action expert at 2 bits}}\\
Full precision & 16 & 76.5 & -- & 74.4 & 78.6 & 82.0 & 81.7 & 76.6 & 65.7 \\
AQLM~\cite{egiazarian2024extreme} & 2.02 & 63.9 & \dn{12.6} & 60.8 & 67.0 & 73.1 & 71.7 & 60.5 & 50.4 \\
QuIP\#~\cite{tseng2024quip} & 2.03 & 66.2 & \dn{10.3} & 64.6 & 67.8 & 73.5 & 70.9 & 68.8 & 51.6 \\
VPTQ~\cite{liu2024vptq} & 2.00 & 68.5 & \dn{8.0} & 65.8 & 71.2 & 78.8 & 74.9 & 68.0 & 52.4 \\
QTIP~\cite{tseng2024qtip} & 2.02 & 72.6 & \dn{3.9} & 71.0 & 74.2 & 79.2 & 77.3 & 71.1 & 62.9 \\
\calrow Closed-loop calibration only (ours) & 2.02 & 73.4 & \dn{3.1} & 70.2 & 76.6 & 81.2 & \textbf{79.7} & 74.2 & 58.5 \\
\oursrow \textbf{\method{}} (ours) & 2.02 & \textbf{74.9} & \textbf{\dn{1.6}} & \textbf{72.8} & \textbf{77.0} & \textbf{82.9} & 77.7 & \textbf{75.8} & \textbf{63.3} \\
\bottomrule
\end{tabular}
\caption{\textbf{LIBERO-Plus against vector quantizers, every linear layer at 2 bits}: success (\%) on 1{,}000 paired instances per model, by perturbation type (500 each) and suite. bpw: bits per weight over the quantized layers, all stored tensors included; bold: best 2-bit result.}
\label{tab:groot}
\end{table*}

\textbf{Spectral robustness view.} Let $H'_\ell$ denote the true second moment at deployment. The deployment reconstruction error satisfies
\begin{equation}
\begin{aligned}
\tr\!\big(\Delta W H'_\ell \Delta W^{\top}\big)\ \le\ &\tr\!\big(\Delta W H_\ell \Delta W^{\top}\big)\\
&+\big\|H'_\ell-H_\ell\big\|_2\,\|\Delta W\|_F^2 ,
\end{aligned}
\label{eq:bound}
\end{equation}
where $\Delta W=\hat W_\ell-W_\ell$. When the shift $H'_\ell-H_\ell$ is unknown, the second term is controlled by diagonal damping ($H_\ell+\lambda I$), which penalizes weight error equally in every direction, including those that deployment never excites. Closed-loop calibration measures the shift instead: $\Hcl_\ell-H^{\mathrm{demo}}_\ell$ concentrates its energy in the subspaces excited by visual variations and on-policy states. The step costs 256 simulated rollouts and a second quantization pass that reuses the quantizer, rotation, storage format and bit budget of the first.

\subsection{Readout-aligned fine-tuning}
Readout-aligned fine-tuning then repairs the representations consumed by the action head. Freezing all trellis codes and rotations, we optimize only the per-layer lookup tables $T_\ell$, row scales $s_\ell$, and normalization weights. Let $o^{q}$ and $o^{\mathrm{FP}}$ denote the attention readouts~\eqref{eq:readout} of the quantized and full-precision policies along the quantized policy's denoising trajectory under identical observations and noise. We minimize
\begin{equation}
\begin{aligned}
\mathcal L =\ & \sum_{\ell,t,h}\big\|W_{O,\ell,h}\big(o^{q}_{\ell,t,h}-o^{\mathrm{FP}}_{\ell,t,h}\big)\big\|^2
+\beta\,\big\|v^{q}-v^{\mathrm{FP}}\big\|^2\\
&+\mu\sum_\ell \frac{\|\hat W_\ell-W_\ell\|_F^2}{\|W_\ell\|_F^2},
\end{aligned}
\label{eq:loss}
\end{equation}
where $W_O$ is the attention output projection, $v$ the predicted velocity over the executed chunk, and the Frobenius anchor keeps decoded weights near full precision ($\mu$ sets it to 10\% of the initial loss).

\textbf{Readout-level alignment}: constraining the projected attention outputs $W_O o$ instead of raw prefix keys and values focuses the limited trainable capacity on the subspace extracted by the action queries. \textbf{Student-side queries}: $o^{q}$ is computed with the quantized action head's own queries, so that gradients flow through the head's lookup tables; this is essential for GR00T, whose 1.1B-parameter head holds half of the quantized weights, whereas freezing the compact head of \pizero{} is a special case, and $\beta$ is set once to balance prefix and head gradient norms. \textbf{Table-scale stability}: each table entry is shared by thousands of weights, so standard codebook learning rates destabilize entire layers. We pair a small cosine-decayed learning rate with a root-mean-square re-normalization of each table after every step, absorbing the factor into $s_\ell$. Fine-tuning sees only the 256 demonstration observations used for calibration: 768 updates on 216 of them, with the checkpoint selected on the other 40.

\begin{table*}[t]
\centering
\small\tabstyle
\setlength{\tabcolsep}{5.2pt}
\begin{tabular}{lcc ccccc c}
\toprule
\hdrrow & \multicolumn{2}{c}{Bits per weight} & \multicolumn{5}{c}{LIBERO success (\%)} & LIBERO-Plus \\
\hdrrow Method & quantized layers & whole model & Spatial & Object & Goal & Long & Avg. & success (\%) \\
\midrule
\grprow\multicolumn{9}{l}{\emph{Reported in~\cite{akbari2026actquant}: vision encoder and language model quantized; projector and action expert kept in higher precision}}\\
Full precision & 16 & 16 & 98.4 & 98.4 & 97.6 & 93.6 & 97.0 & -- \\
QVLA~\cite{xu2026qvla} & 2.5 & 5.3 & 93.0 & 97.6 & 71.0 & 55.0 & 79.2 & -- \\
QVLA~\cite{xu2026qvla} & 2.0 & 4.7 & 0.0 & 0.0 & 0.0 & 0.0 & 0.0 & -- \\
ActQuant~\cite{akbari2026actquant} & 3.0 & 5.6 & 98.2 & 98.8 & 95.0 & 87.2 & 94.8 & -- \\
ActQuant~\cite{akbari2026actquant} & 2.5 & 5.2 & 96.6 & 98.4 & 74.0 & 73.6 & 85.7 & -- \\
ActQuant~\cite{akbari2026actquant} & 2.0 & 4.8 & 61.2 & 80.0 & 25.0 & 25.0 & 48.0 & -- \\
\midrule
\grprow\multicolumn{9}{l}{\emph{Our evaluation, 2 bits: all 414 linear layers of backbone and action expert quantized}}\\
Full precision & 16 & 16 & 98.2 & 98.8 & 96.8 & 91.8 & 96.4 & 76.5 \\
AWQ~\cite{lin2024awq} & 2.15 & 3.02 & 80.2 & 68.0 & 44.4 & 13.6 & 51.5 & 15.4 \\
GPTQ~\cite{frantar2022gptq} & 2.14 & 3.02 & 89.6 & 97.2 & 47.6 & 45.6 & 70.0 & 39.3 \\
AQLM~\cite{egiazarian2024extreme} & 2.02 & 2.92 & 98.2 & 99.0 & 95.4 & 90.4 & 95.8 & 63.9 \\
QuIP\#~\cite{tseng2024quip} & 2.03 & 2.93 & 98.0 & 98.0 & 97.4 & 91.6 & 96.3 & 66.2 \\
VPTQ~\cite{liu2024vptq} & 2.00 & 2.91 & 98.4 & 98.6 & 98.0 & 93.0 & 97.0 & 68.5 \\
QTIP~\cite{tseng2024qtip} & 2.02 & 2.93 & 98.6 & 99.6 & 97.8 & 94.2 & 97.6 & 72.6 \\
\calrow Closed-loop calibration only (ours) & 2.02 & 2.92 & 98.6 & 99.6 & 98.2 & 96.0 & 98.1 & 73.4 \\
\oursrow \textbf{\method{}} (ours) & 2.02 & 2.92 & 98.4 & 99.2 & 97.8 & 94.8 & 97.6 & \textbf{74.9} \\
\midrule
\grprow\multicolumn{9}{l}{\emph{Our evaluation, 1 bit: the same 414 layers}}\\
VPTQ~\cite{liu2024vptq} & 1.00 & 2.10 & \multicolumn{5}{c}{fails} & 0.0 \\
QuIP\#~\cite{tseng2024quip} & 1.02 & 2.11 & 55.6 & 25.2 & 18.4 & 0.0 & 24.8 & 13.8 \\
AQLM~\cite{egiazarian2024extreme} & 1.01 & 2.11 & 70.0 & 29.6 & 20.6 & 1.0 & 30.3 & 18.1 \\
QTIP~\cite{tseng2024qtip} & 1.02 & 2.12 & 92.0 & 92.2 & 50.0 & 28.2 & 65.6 & 39.9 \\
\calrow Closed-loop calibration only (ours) & 1.02 & 2.11 & 95.0 & 98.6 & 58.2 & 43.6 & 73.9 & 45.0 \\
\oursrow \textbf{\method{}} (ours) & 1.02 & 2.11 & \textbf{95.2} & \textbf{99.0} & \textbf{69.4} & \textbf{56.4} & \textbf{80.0} & \textbf{49.8} \\
\bottomrule
\end{tabular}
\caption{\textbf{\pizero{} on LIBERO and LIBERO-Plus at 2 bits and 1 bit.} Top: results reported by ActQuant~\cite{akbari2026actquant}, whose bit budget covers the vision encoder and language model only (different protocol, shown for reference). Middle and bottom: our evaluation with every linear layer of backbone and action expert quantized (2{,}000 LIBERO episodes; 1{,}000 paired LIBERO-Plus instances). }
\label{tab:pi05}
\end{table*}

\subsection{Deployed format and inference}
The stored model is a packed stream of 2-bit codes with one 512-entry table and one scale vector per layer, 2.02 bits per weight over all linear layers, unchanged by both steps. Each weight pair is decoded from a fixed 16-bit window of the stream, so all pairs decode independently, with one shift and one table lookup, inside the GPU kernel that uses them. With 8-bit tables and 8-bit activations taken in the rotated space, a layer becomes an integer matrix product followed by one scale per row. One file runs in two modes: decoding once at load gives the highest speed, and decoding on the fly keeps only the 2-bit stream in GPU memory. Both modes reproduce the reference arithmetic bit for bit (\cref{sec:deploy}).
%=====================================================================

%=====================================================================
\section{Experiments}
\label{sec:exp}

\subsection{Setup}
\textbf{Models and benchmarks.} We quantize every linear layer of GR00T~N1.7~\cite{bjorck2025gr00t} (216 in the vision-language backbone, 216 in the 1.1B-parameter diffusion-transformer head) and of \pizero{}~\cite{intelligence2025pi_} (288 in the backbone, 126 in the 0.3B action expert). We evaluate on LIBERO~\cite{liu2023libero}, on a fixed set of 1{,}000 perturbed LIBERO-Plus instances~\cite{fei2025libero}, 500 with changed camera viewpoints and 500 with changed robot initial states across the four suites, and on a real AgileX Piper arm. All methods run on identical instances, and we report the mean over three noise seeds. Results on the SimplerEnv Bridge tasks~\cite{li2024evaluating} are in the supplementary material.

\textbf{Baselines.} AQLM~\cite{egiazarian2024extreme}, VPTQ~\cite{liu2024vptq}, QuIP\#~\cite{tseng2024quip}, QTIP~\cite{tseng2024qtip} and the scalar quantizers GPTQ~\cite{frantar2022gptq} and AWQ~\cite{lin2024awq}, run layer-wise from their official toolkits, without end-to-end fine-tuning, on the same 256 calibration observations as \method{}, so our calibration-only rows are the training-free comparison; bits per weight count every stored tensor. On \pizero{} we also list the results reported by ActQuant~\cite{akbari2026actquant} and QVLA~\cite{xu2026qvla}. Hyperparameters and compute cost are in the supplementary material.

\begin{figure*}[t]
\centering
\includegraphics[width=\linewidth]{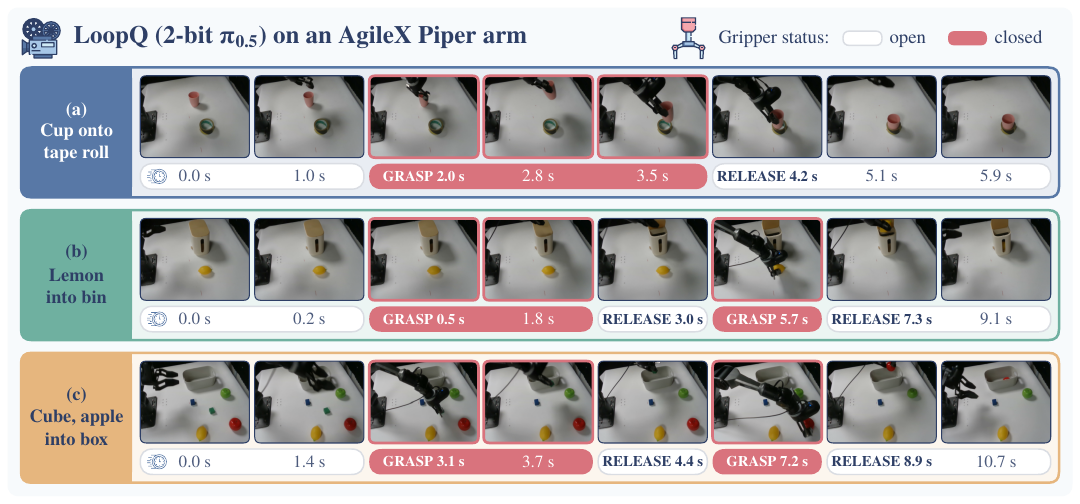}
\caption{\textbf{\method{} (2-bit \pizero{}) on an AgileX Piper arm.} One successful episode of each task: (a)~stacking a cup onto a tape roll, (b)~placing a lemon into a lidded bin, and (c)~moving a cube and an apple into a box in clutter. The ribbon under the frames gives their time stamps and the gripper state. Videos are in the supplementary material.}
\label{fig:real}
\end{figure*}

\subsection{Simulation results}

\textbf{GR00T~N1.7} (\cref{tab:groot}, top). The strongest baseline, QTIP, loses 5.9 points and the other vector quantizers 9 to 22. In QTIP's storage format, closed-loop calibration adds 1.6 points without any training, exactly where the analysis predicts: +3.4 under viewpoint changes with the initial-state half preserved, and +5.0 on LIBERO-Goal, whose tasks share scenes and differ only in the instructed goal. Readout-aligned fine-tuning adds a further 2.9. \method{} reaches 60.4\%, 4.5 points above the best baseline and within 1.4 of full precision, recovering three quarters of what 2-bit quantization costs at the same bit budget and decoder; under initial-state perturbations and on the Spatial suite it is on par with full precision.

\textbf{\pizero{}} (\cref{tab:pi05}; \cref{tab:groot}, bottom). ActQuant~\cite{akbari2026actquant}, the strongest published VLA quantizer below four bits, keeps the action expert and projector in higher precision, so its nominal 2.0 bits correspond to 4.8 bits over the whole policy; our rows quantize the action expert as well and reach 2.9 bits. At this harder operating point, where ActQuant retains 48.0\% on LIBERO, the vector quantizers stay at the level of full precision and our two rows reach 98.1\% and 97.6\% against 96.4\%. Since unperturbed LIBERO is saturated at 2 bits, the methods separate on LIBERO-Plus, where the baselines lose 4 to 13 points, closed-loop calibration alone surpasses all of them, and \method{} narrows the gap to 1.6 points. It leads under both perturbation types (72.8\% vs.\ 71.0\% for QTIP under viewpoint changes, 77.0\% vs.\ 74.2\% under initial-state perturbations) and on three of the four suites, exceeding full precision on Spatial.

\textbf{One bit per weight} (\cref{tab:pi05}, bottom). The trellis coder changes rate by changing the bits consumed per step, so the same pipeline runs at 1 bit unchanged. Existing quantizers do not survive this rate: VPTQ fails, QuIP\# and AQLM fall to 25\% and 30\% on LIBERO, and QTIP retains 65.6\%. \method{} remains a working policy at 80.0\%, within three points of full precision on the Spatial and Object suites and 14.4 points above the best baseline (9.9 on LIBERO-Plus), and closed-loop calibration alone exceeds every baseline. Readout alignment matters more the less precision is left: it lifts GR00T on LIBERO-Plus from 8.1\% to 29.2\%.

\subsection{Ablations}
\label{sec:ablation}

\textbf{Calibration Hessian.} Without fine-tuning on GR00T, adding on-policy states collected from a single start pose to $H^{\mathrm{demo}}$ changes LIBERO-Plus success from 55.9\% to 55.3\%: robustness to viewpoint rises by 3.6 points, but robustness to the initial state falls by 4.8, because one pose covers fewer states than demonstrations. Starting rollouts from the benchmark's initial-state distribution, as in $\Hcl$, keeps the gain and removes the loss (57.5\%). What the calibration statistics cover is what the 2-bit model is robust to.

\textbf{Complementarity.} On held-out observations, the action error of the 2-bit policy relative to full precision falls from $5.70\times10^{-3}$ for the base to $5.15\times10^{-3}$ with closed-loop calibration and to $4.59\times10^{-3}$ with fine-tuning on top, 20\% in total, and the readout error falls by 13\%: the two steps address different causes, and their effects add. The supplementary material ablates the sources and the weighting of $\Hcl$, a second fixed-point iteration, and the design choices of the fine-tuning objective.

\begin{table}[t]
\centering
\small\tabstyle
\setlength{\tabcolsep}{2.25pt}
\begin{tabular}{l cccc cccc}
\toprule
\hdrrow & \multicolumn{4}{c}{Successes / 25 trials} & \multicolumn{4}{c}{RTX 6000 Ada} \\
\hdrrow & (a) & (b) & (c) & Avg. & Disk & Mem. & Prefill & Chunk \\
\hdrrow \pizero{} policy & & & & (\%) & (GB) & (GB) & (ms) & (ms) \\
\midrule
Full precision & 25 & 21 & 13 & 78.7 & 6.47 & 6.65 & 51.8 & 81.6 \\
\oursrow \textbf{\method{}} (2-bit) & 25 & 21 & 12 & 77.3 & \textbf{1.58} & \textbf{3.72} & \textbf{38.2} & \textbf{74.6} \\
\bottomrule
\end{tabular}
\caption{\textbf{Real-world deployment of \pizero{}} on an AgileX Piper arm (tasks as in \cref{fig:real}) and inference on the RTX 6000 Ada that drives it, with integer kernels and 8-bit activations.}
\label{tab:real}
\end{table}

\subsection{Real-world deployment}
\label{sec:deploy}

\textbf{Real robot} (\cref{fig:real}, \cref{tab:real}). We deploy \pizero{} on an AgileX Piper arm and compare the 2-bit \method{} policy with its full-precision counterpart on three tasks, 25 trials each: stacking a cup onto a tape roll, which demands precise placement; putting a lemon into a lidded bin, a two-stage task; and moving a cube and an apple into a box in clutter. Its on-policy states are collected by rolling out the 2-bit policy on the real arm. The 2-bit policy succeeds in 58 of 75 trials against 59, matching full precision on the first two tasks and staying within one trial on the third. On the RTX 6000 Ada that drives the arm it is also the faster policy: an action chunk takes 74.6\,ms against 81.6\,ms and the prefill 38.2\,ms against 51.8\,ms, while the weights shrink from 6.47 to 1.58\,GB and peak GPU memory from 6.65 to 3.72\,GB. The speed comes from custom CUDA kernels that decode the weights once at load and run each layer as an 8-bit integer product, with one shared rotation and quantization per input tensor.

\textbf{Embedded platform.} On an NVIDIA Jetson AGX Thor, an embedded computer built for robots, the policy with 2-bit weights and 8-bit activations stores 1.58\,GB on disk instead of the 6.47\,GB of full precision ($4.1\times$ smaller) and holds 3.72\,GB of memory instead of 6.65\,GB at inference (44\% less). Kernel design and all operating points are in the supplementary material.

%=====================================================================
\section{Conclusion}
\label{sec:conclusion}
A quantizer treats a policy as weights to approximate, but a policy is a closed loop: it chooses its own inputs and acts on what its action head reads. \method{} lets calibration follow the quantized policy and the objective follow the readout, and two bits per weight then keep an entire VLA within two points of full precision under perturbation and within one success of it on a real arm, while one bit still yields a working policy. The lesson reaches beyond quantization: an embodied model should be compressed on the distribution it induces and for the signals it consumes.

% \clearpage
{
    \small
    \bibliographystyle{ieeenat_fullname}
    \bibliography{main}
}

%=====================================================================
% Supplementary material: paste after the bibliography, before \end{document}.
% Uses the table macros of the main text (\tabstyle, \hdrrow, \grprow, \calrow, \oursrow).
%=====================================================================
\clearpage
\setcounter{page}{1}
\backrefsetup{disable}% citations of the supplement leave the back-references of the main text as they are
\raggedbottom
\appendix
\renewcommand{\thetable}{S\arabic{table}}
\renewcommand{\theHtable}{S\arabic{table}}
\setcounter{table}{0}
\renewcommand{\thefigure}{S\arabic{figure}}
\renewcommand{\theHfigure}{S\arabic{figure}}
\setcounter{figure}{0}
% title of the supplement with Table S1 across both columns
\twocolumn[{%
\centering
{\Large\textbf{\thetitle}\\ \vspace{0.5em}Supplementary Material\par}
\vspace{1.2em}
{\captionsetup{type=table}%
{\small\tabstyle
\setlength{\tabcolsep}{8.6pt}
\begin{tabular}{l c cc cccc}
\toprule
\hdrrow & & \multicolumn{2}{c}{By perturbation} & \multicolumn{4}{c}{By suite} \\
\hdrrow Variant & All & Camera & Init.\ state & Spatial & Object & Goal & Long \\
\midrule
\grprow\multicolumn{8}{l}{\emph{Calibration sources added to $H^{\mathrm{demo}}$ (no fine-tuning)}}\\
None (demonstrations only) & 55.9 & 54.4 & \textbf{57.4} & \textbf{71.4} & 48.6 & 60.2 & 43.5 \\
On-policy states & 54.8 & 56.4 & 53.2 & 66.9 & 47.0 & 61.3 & 44.0 \\
Lighting and crop & 55.4 & 56.6 & 54.2 & 66.5 & 49.8 & 58.6 & \textbf{46.8} \\
\calrow Lighting, crop and on-policy states ($\Hcl$) & \textbf{57.5} & \textbf{57.8} & 57.2 & 71.0 & \textbf{50.6} & \textbf{65.2} & 43.1 \\
\midrule
\grprow\multicolumn{8}{l}{\emph{Weighting of the three sources (no fine-tuning)}}\\
Plain sum, no trace normalization & 54.4 & 54.0 & 54.8 & 67.8 & 46.2 & 62.1 & 41.5 \\
Trace-normalized, total weight $\times 0.5$ & 55.9 & 56.6 & 55.2 & 66.5 & 46.2 & 64.5 & \textbf{46.4} \\
Trace-normalized, total weight $\times 2$ & 55.8 & \textbf{58.6} & 53.0 & 67.8 & 48.6 & 63.7 & 43.1 \\
\calrow Trace-normalized, total weight 1 ($\Hcl$) & \textbf{57.5} & 57.8 & \textbf{57.2} & \textbf{71.0} & \textbf{50.6} & \textbf{65.2} & 43.1 \\
\midrule
\grprow\multicolumn{8}{l}{\emph{Fixed-point iterations (no fine-tuning)}}\\
\calrow One ($\Hcl$) & \textbf{57.5} & 57.8 & \textbf{57.2} & \textbf{71.0} & \textbf{50.6} & \textbf{65.2} & 43.1 \\
Two & 56.7 & \textbf{59.4} & 54.0 & 70.2 & 49.4 & 62.1 & \textbf{45.2} \\
\midrule
\grprow\multicolumn{8}{l}{\emph{Fine-tuning objective, starting from the closed-loop-calibrated model}}\\
Readout term only ($\beta=0$) & 58.6 & 61.0 & 56.2 & 71.8 & \textbf{50.6} & 63.7 & 48.4 \\
Velocity term only & 57.9 & 61.0 & 54.8 & 70.6 & 47.4 & 67.6 & 46.0 \\
Teacher-side queries & 57.9 & 59.8 & 56.0 & 74.7 & 47.0 & 65.2 & 44.8 \\
Without table re-normalization & 59.2 & 62.0 & 56.4 & 72.7 & 49.4 & 65.2 & \textbf{49.6} \\
\oursrow \textbf{\method{}} (\cref{tab:groot}) & \textbf{60.4} & \textbf{62.2} & \textbf{58.6} & \textbf{75.1} & 49.0 & \textbf{68.4} & 49.2 \\
\bottomrule
\end{tabular}\par}
\vspace{0.6em}
\caption{\textbf{Ablations of \method{} on GR00T~N1.7 at 2 bits}: LIBERO-Plus success (\%) on the 1{,}000 instances of \cref{tab:groot}. Green rows are the default $\Hcl$ and the blue row is \method{}; bold: best of each group.}
\label{tab:supp_abl}}
\vspace{1.6em}
}]

This supplement provides comprehensive ablations of \method{} (\cref{sec:supp_abl}), additional simulation and real robot results (\cref{sec:supp_results}), full implementation details (\cref{sec:supp_impl}), and the design and benchmarking of our inference kernels (\cref{sec:supp_deploy}). Accompanying videos document the real robot evaluation episodes.

%---------------------------------------------------------------------
\section{Ablations of \method{}}
\label{sec:supp_abl}

All ablations evaluate GR00T~N1.7 with every linear layer quantized to 2 bits across the 1{,}000 LIBERO-Plus instances of \cref{tab:groot}. Each group in \cref{tab:supp_abl} isolates a single design choice while keeping the remaining pipeline at its default configuration.

\begin{figure*}[t]
\centering
\includegraphics[width=\linewidth]{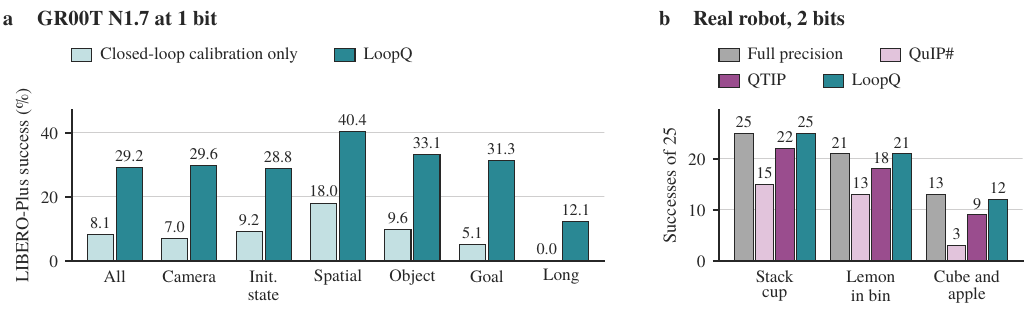}
\caption{\textbf{Additional results.} \textbf{(a)}~GR00T~N1.7 at 1 bit per weight on the 1{,}000 LIBERO-Plus instances of \cref{tab:groot}. \textbf{(b)}~\pizero{} at 2 bits on the AgileX Piper arm: successes in 25 trials on the three tasks of \cref{fig:real}.}
\label{fig:supp_results}
\end{figure*}

\begin{table*}[t]
\centering
\footnotesize\tabstyle\renewcommand{\arraystretch}{1.1}
\setlength{\tabcolsep}{6pt}
\newcommand{\both}[1]{\multicolumn{2}{p{\dimexpr0.68\linewidth+2\tabcolsep\relax}}{#1}}
\begin{tabular}{l p{0.34\linewidth} p{0.34\linewidth}}
\toprule
\hdrrow Setting & GR00T~N1.7 & \pizero{} \\
\midrule
\grprow\multicolumn{3}{l}{\emph{Format}}\\
Trellis code & \both{bitshift, 16-bit window, 2 weights per step, 4 bits per step (2 bits per step at 1 bit per weight)} \\
Lookup table and scale & \both{512 entries of 2 values and one scale per output row, 16-bit floats (8-bit tables at inference)} \\
Search & \both{Viterbi with block-wise error feedback, one pass; damping 0.01 of the mean diagonal of $H_\ell$} \\
Bits per weight, linear layers & 2.018 (2 bits), 1.018 (1 bit) & 2.016 (2 bits), 1.016 (1 bit) \\
\midrule
\grprow\multicolumn{3}{l}{\emph{Closed-loop calibration}}\\
Lighting & \both{2 copies per observation: brightness and contrast in $[0.8,1.2]$, hue $\pm5^\circ$, Gaussian blur with $\sigma\in[0.5,1.0]$} \\
Crop & \both{1 copy per observation: square crop of 90--95\% of the area, resized to the input resolution} \\
On-policy states & \both{256 rollouts from the official initial states of each task; one state from the first 40\% of the policy calls} \\
\midrule
\grprow\multicolumn{3}{l}{\emph{Readout-aligned fine-tuning}}\\
Trainable parameters & tables and row scales of all 432 layers, normalization weights & tables and row scales of the 288 backbone layers, normalization weights; action expert frozen \\
Velocity weight $\beta$ & matches the gradient norms of the two terms at the first update & 0.1 \\
Frobenius anchor $\mu$ & \both{sets the anchor to 10\% of the initial loss} \\
Schedule & \both{768 updates of 8 samples; Adam; 5\% linear warm-up, cosine decay; gradient clipping at 1.0} \\
Learning rates & \both{$2\times10^{-5}$ for tables and normalization weights, $10^{-4}$ for the logarithm of the row scales} \\
Table re-normalization & \both{each table rescaled to its initial root mean square after every update, factor absorbed into $s_\ell$} \\
Exported weights & \both{exponential moving average (0.99); checkpoint with the lowest held-out readout loss, evaluated every 64 updates} \\
Data & \both{216 calibration observations for training, 40 held out} \\
\bottomrule
\end{tabular}
\caption{\textbf{Hyperparameters of \method{}.} Entries spanning both columns are shared by the two models.}
\label{tab:supp_hparams}
\end{table*}

\textbf{Calibration sources.} Combining all three closed-loop sources in $\Hcl$ delivers the highest overall success rate (57.5\%, $+1.6$ points over demonstration calibration alone) and exhibits strong synergy across perturbation types. Adding on-policy states or visual augmentations in isolation improves camera robustness by roughly two points (56.4\% and 56.6\% vs.\ 54.4\%) but reduces initial state generalization (53.2\% and 54.2\% vs.\ 57.4\%). Integrating all three sources resolves this tension cleanly: camera robustness jumps by 3.4 points to 57.8\%, Object and Goal success rise sharply to 50.6\% ($+2.0$) and 65.2\% ($+5.0$), and initial state robustness is preserved at 57.2\%. All three sources are therefore mutually complementary and jointly necessary to unlock the full calibration gain.

\textbf{Source weighting.} Equal trace normalization with a total weight of 1 decisively outperforms all alternative schemes, achieving 57.5\% overall while leading on initial state robustness (57.2\%) as well as the Spatial, Object, and Goal suites. Because the raw second moments of each added source carry roughly the same energy as the demonstrations, a plain unnormalized sum overweights the auxiliary sources threefold and degrades accuracy by 3.1 points to 54.4\%. Trace normalization eliminates this scale imbalance, and unit total weight strikes the optimal balance between demonstration fidelity and closed-loop coverage, outperforming both halved (55.9\%, $-1.6$) and doubled (55.8\%, $-1.7$) weights.

\textbf{Fixed-point iterations.} A single closed-loop iteration already converges closely to the fixed point while achieving higher overall accuracy (57.5\% vs.\ 56.7\% for two iterations) at half the rollout cost. Recollecting 256 rollouts from the $\Hcl$ calibrated policy shifts the trace normalized second moments by a mean relative Frobenius distance of only 0.014 in the vision encoder, 0.010 in the language model, and 0.054 in the action head, an order of magnitude smaller than the first iteration's shift away from the demonstrations (0.140, 0.083, and 0.268). One iteration thus captures virtually the entire distribution shift without incurring additional rollout overhead.

\textbf{Fine-tuning objective.} Starting from the closed-loop calibrated model (57.5\%), the full \method{} objective lifts overall success by 2.9 points to 60.4\% (\cref{tab:groot}), decisively outperforming every ablated variant on both camera (62.2\%) and initial state (58.6\%) perturbations as well as the Spatial (75.1\%) and Goal (68.4\%) suites. Each component contributes directly to this gain. The readout term alone reaches 58.6\% and cuts the held-out readout loss from 7{,}998 to 6{,}464, whereas the velocity term alone yields only 57.9\% and leaves the readout loss nearly unchanged (7{,}757), confirming that explicit readout alignment is what repairs the quantized cross attention interface. Coupling both terms adds a further $+1.8$ points over readout training alone. Evaluating the readout with student queries rather than teacher queries is likewise crucial ($+2.5$ points, 60.4\% vs.\ 57.9\%) because it aligns the representation under the student's actual inference distribution, and table re-normalization contributes another $+1.2$ points (60.4\% vs.\ 59.2\%) by keeping codebook scales properly conditioned. Finally, aligning raw prefix keys and values in place of $W_O o$ starts from a $30\times$ larger loss and diverges with non-finite gradients after 245 of 768 updates under identical optimizer settings, underscoring the superior conditioning and stability of our readout formulation.

\begin{figure*}[t]
\centering
\includegraphics[width=\linewidth]{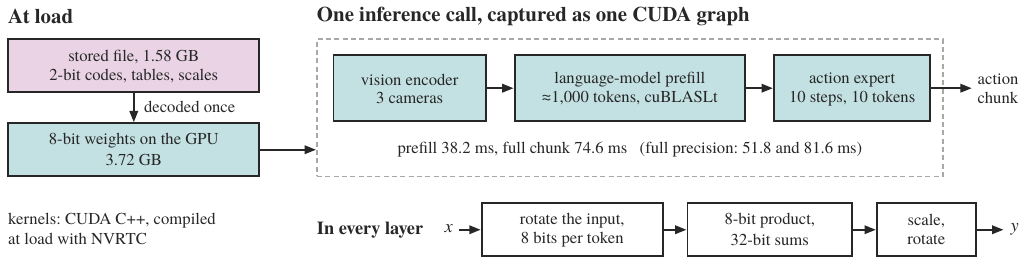}
\caption{\textbf{Deployment of the 2-bit \pizero{} with 8-bit activations on an RTX 6000 Ada.} The stored file is decoded once at load into 8-bit weights. One CUDA graph then runs each inference call, and every quantized layer executes the three steps at the bottom.}
\label{fig:supp_deploy}
\end{figure*}

%---------------------------------------------------------------------
\section{Additional results}
\label{sec:supp_results}

\textbf{GR00T~N1.7 at 1 bit.} \Cref{fig:supp_results}a breaks down the extreme 1-bit result from the main text. Readout-aligned fine-tuning more than triples the success rate of the calibrated 1-bit policy, jumping by 21.1 points from 8.1\% to 29.2\%, with consistent and substantial gains under both perturbation types and across all four suites.

\textbf{Real robot against 2-bit baselines.} \Cref{fig:supp_results}b compares \method{} against two strong 2-bit baselines on the physical AgileX Piper arm (\cref{tab:real}). \method{} achieves 58 of 75 successes, virtually matching full precision (59/75), while outperforming QTIP (49/75) by 9 trials and QuIP\# (31/75) by 27 trials. This advantage widens markedly on the hardest cluttered two-object task, where \method{} succeeds 12 times (vs.\ 13 for full precision) while QTIP and QuIP\# fall to 9 and 3 successes, respectively. The accompanying videos show the 2-bit \method{} policy executing all three tasks of \cref{fig:real}.

\begin{table}[t]
\centering
\small\tabstyle
\setlength{\tabcolsep}{5pt}
\begin{tabular}{l cc}
\toprule
\hdrrow Method & Successes & (\%) \\
\midrule
Full precision & 507 & 60.4 \\
AWQ~\cite{lin2024awq} & 0$^{\ast}$ & 0.0 \\
QuIP\#~\cite{tseng2024quip} & 533 & 63.5 \\
Trellis, demonstration calibration & 514 & 61.2 \\
\calrow Closed-loop calibration only (ours) & 512 & 61.0 \\
\oursrow \textbf{\method{}} (ours) & \textbf{536} & \textbf{63.8} \\
\bottomrule
\end{tabular}
\caption{\textbf{SimplerEnv Bridge with GR00T~N1.7 at 2 bits}: successes out of 840 episodes (seven WidowX tasks, 120 seeded episodes each), all rows on one RTX 6000 Ada. $^{\ast}$Stopped after 509 episodes.}
\label{tab:supp_bridge}
\end{table}

\textbf{SimplerEnv Bridge.} We apply the same 2-bit quantization pipeline to the GR00T~N1.7 Bridge checkpoint, calibrating on 256 Bridge demonstration observations and, for the closed-loop rows, 256 SimplerEnv rollouts. Each policy is evaluated across all seven WidowX tasks for 120 seeded episodes (\cref{tab:supp_bridge}). While AWQ collapses completely (0.0\%), \method{} achieves the highest overall success rate at 63.8\% (536/840), outperforming full precision by 3.4 points (60.4\%), trellis quantization with demonstration calibration by 2.6 points (61.2\%), closed-loop calibration alone by 2.8 points (61.0\%), and QuIP\# (63.5\%). Because Bridge evaluates standard layouts without external perturbations, all 2-bit trellis variants remain at or above full precision, and readout-aligned fine-tuning still delivers a clear gain on top of calibration. These results confirm that \method{} transfers cleanly across robot embodiments and benchmarks without sacrificing nominal accuracy.

%---------------------------------------------------------------------
\section{Implementation details}
\label{sec:supp_impl}

\textbf{Hyperparameters.} \Cref{tab:supp_hparams} summarizes the complete settings for both models. The 1-bit models share the same configuration while consuming 2 bits per trellis step.

\textbf{Baselines.} All baselines follow their official implementations and use the same 256 demonstration observations as \method{}. AQLM~\cite{egiazarian2024extreme} uses 2 codebooks of 8 bits over groups of 8 weights. VPTQ~\cite{liu2024vptq} uses vectors of length 4 with 256 centroids. QuIP\#~\cite{tseng2024quip} uses the E8P12 codebook with its default 10 tuning sweeps. QTIP~\cite{tseng2024qtip} uses the bitshift trellis of \cref{tab:supp_hparams}. GPTQ~\cite{frantar2022gptq} and AWQ~\cite{lin2024awq} quantize to 2 bits with group size 128.

\textbf{Compute cost.} Every stage runs on a single GPU (RTX 6000 Ada or RTX A6000). For GR00T~N1.7, collecting the 256 on-policy states takes 32 minutes and accumulating the second moments of the four sources takes about 15 minutes. Quantization takes 2.2 hours for the GR00T~N1.7 backbone, 1.9 to 2.8 hours per suite for the action head, and 3.9 hours for all of \pizero{}. Readout-aligned fine-tuning takes 18.7 hours for GR00T~N1.7 and 3.3 hours for \pizero{}.

%---------------------------------------------------------------------
\section{Inference kernels and operating points}
\label{sec:supp_deploy}

\textbf{Kernel design.} Each quantized layer computes $y=S_vH\,\mathrm{diag}(s_\ell)\,Q\,HS_ux$, where $Q$ holds the decoded table values, $H$ is the Hadamard transform, and $S_u,S_v$ are the stored sign vectors. Our CUDA kernels execute this in four steps: rotating and quantizing the input to 8 bits per token, decoding all weight pairs in parallel from their 16-bit windows, computing an 8-bit integer product with 32-bit accumulation, and applying the token and row scales together with the output rotation. Projections reading the same input tensor (query, key, and value; the two MLP input projections) share a single input rotation, eliminating 162 of the 414 input rotations exactly. The 8-bit tables are derived from the stored tables at load with a maximum relative weight error of only $5.9\times10^{-5}$.

\begin{table}[t]
\centering
\small\tabstyle
\setlength{\tabcolsep}{3.2pt}
\begin{tabular}{l cccc}
\toprule
\hdrrow & Disk & Mem. & Prefill & Chunk \\
\hdrrow Operating point & (GB) & (GB) & (ms) & (ms) \\
\midrule
Full precision & 6.47 & 6.65 & 51.8 & 81.6 \\
\grprow\multicolumn{5}{l}{\emph{2-bit weights, floating-point activations}}\\
Decoded at load & 1.58 & 6.16 & 52.4 & 82.1 \\
Decoded on the fly & 1.58 & 1.99 & 86.5 & 133.1 \\
\grprow\multicolumn{5}{l}{\emph{2-bit weights, 8-bit activations, integer kernels}}\\
\oursrow Decoded at load & 1.58 & 3.72 & \textbf{38.2} & \textbf{74.6} \\
\oursrow Decoded on the fly & 1.58 & \textbf{1.92} & 40.6 & 79.8 \\
\bottomrule
\end{tabular}
\caption{\textbf{Operating points of the 2-bit \pizero{}} on an RTX 6000 Ada: median of 200 runs at batch size 1 with 10 denoising steps. Memory is the peak GPU allocation after load.}
\label{tab:supp_ops}
\end{table}

\textbf{Deployment on the RTX 6000 Ada} (\cref{fig:supp_deploy}). The policy runs in PyTorch~2.7 with CUDA~12.6 on a single RTX 6000 Ada. Our CUDA C++ kernels are compiled at load via NVRTC and launched on PyTorch's native stream, enabling a single CUDA graph to capture the entire inference call. The prefill stage (about 1{,}000 tokens) executes its integer GEMMs on cuBLASLt, while the 10-token action expert uses a custom kernel built on 8-bit dot-product instructions. The speedup over full precision is driven by prefill, where matrix products drop by $3.1\times$ from 34\,ms in 16-bit floating point to 11\,ms.

\textbf{Operating points.} \Cref{tab:supp_ops} benchmarks all operating points of the stored 2-bit checkpoint. Both integer kernel modes beat full precision in latency while dramatically reducing GPU memory. Decoding at load (reported in \cref{tab:real}) cuts chunk latency from 81.6\,ms to 74.6\,ms and memory by 44\% to 3.72\,GB. Decoding on the fly retains only the 2-bit stream on the GPU, slashing peak memory by 71\% to 1.92\,GB (29\% of full precision) while still running faster than full precision (79.8\,ms vs.\ 81.6\,ms). Both modes match a PyTorch reference implementation bit for bit across all 414 layers and in the final actions of all 40 held-out observations.

\end{document}